\documentclass{article}

\usepackage{microtype}
\usepackage{graphicx}
\usepackage{subfigure}
\usepackage{booktabs} 

\usepackage{hyperref}
\usepackage[hyphenbreaks]{breakurl}

\usepackage[accepted]{icml2023}

\usepackage{amsmath}
\usepackage{amssymb}
\usepackage{mathtools}
\usepackage{amsthm}
\usepackage{footmisc}
\usepackage{makecell}

\usepackage[capitalize,noabbrev]{cleveref}

\theoremstyle{plain}

\theoremstyle{definition}

\theoremstyle{remark}

\usepackage[textsize=tiny]{todonotes}

\icmltitlerunning{Monotonic Location Attention}

\begin{document}

\twocolumn[
\icmltitle{Monotonic Location Attention for Length Generalization}



\icmlsetsymbol{equal}{*}

\begin{icmlauthorlist}
\icmlauthor{Jishnu Ray Chowdhury}{yyy}
\icmlauthor{Cornelia Caragea}{yyy}
\end{icmlauthorlist}

\icmlaffiliation{yyy}{Computer Science, University of Illinois Chicago}

\icmlcorrespondingauthor{Jishnu Ray Chowdhury}{jraych2@uic.edu}
\icmlcorrespondingauthor{Cornelia Caragea}{cornelia@uic.edu}

\icmlkeywords{Machine Learning, ICML}

\vskip 0.3in
]



\printAffiliationsAndNotice{}  

\begin{abstract}
We explore different ways to utilize position-based cross-attention in seq2seq networks to enable length generalization in algorithmic tasks. We show that a simple approach of interpolating the original and reversed encoded representations combined with relative attention allows near-perfect length generalization for both forward and reverse lookup tasks or copy tasks that had been generally hard to tackle. We also devise harder diagnostic tasks where the relative distance of the ideal attention position varies with timestep. In such settings, the simple interpolation trick with relative attention is not sufficient. We introduce novel variants of location attention building on top of \citet{dubois-etal-2020-location} to address the new diagnostic tasks. We also show the benefits of our approaches for length generalization in SCAN \cite{lake2018generalization} and CFQ \cite{keysers2020measuring}. Our code is available on GitHub\footnote{\url{https://github.com/JRC1995/MonotonicLocationAttention}}.
\end{abstract}

\section{Introduction}
Neural seq2seq \cite{sutskever2014seq2seq} is a powerful generic framework for the task of transforming an input sequence of arbitrary length into an output sequence of arbitrary length. Although seq2seq models can perform impressively in a great variety of tasks \cite{raffel2020exploring, lewis-etal-2020-bart}, they can still struggle in out-of-distribution generalization (e.g., systematic generalization or length generalization), and sometimes, even in simple algorithmic tasks \cite{kim2022uncontrolled, dubois-etal-2020-location, dehghani2018universal, lake2018generalization, liska2018memorize}. Even after extensive pre-training, neural models can show mixed results in such forms of generalization \cite{kim2022uncontrolled, anil2022exploring}. 

In this paper, we focus on length generalization, i.e., the ability of a model to generalize to sequences of unseen (and typically higher) lengths. Particularly, we concentrate on enhancing the interlayer attention mechanism in seq2seq encoder-decoder models for improved length generalization. Similar to \citet{csordas2022the}, we take a {\em bottom up} approach to model development and explore the effects of different strategies of increasing complexities on a range of controlled synthetic probing tasks---each targeting a narrowly defined model behavior or phenomenon---to investigate which strategy works and to what extent, and why does or does not work, and thus, each task precisely pinpointing their capabilities as well as their limitations. Such thorough investigation in a natural language domain can be difficult for at least the following reasons: (1) it can be hard to isolate the exact reasons of failure in natural language due to its complexities and diversity; (2) often there can be exploitable heuristics like emphasis on recency that may improve the overall length generalization performance but preserve systematic issues leading to failures in cases where the heuristics do not apply. Such failures may not be reflected in the overall evaluation if the heuristics apply in the majority of the samples. Besides these factors, the simple synthetic tests that we consider here can be still fairly challenging for neural models. We believe they offer an initial step toward the design of more general-purpose models. 

\begin{table*}[t]
\small
\centering
\def\arraystretch{1.2}
\begin{tabular}{  l | l | l} 
\hline
\textbf{Task} & \textbf{Input} & \textbf{Output}\\
\hline
Copy & $4$ $7$ $9$ $8$ & $4$ $7$ $9$ $8$\\
Reverse Copy & $4$ $7$ $9$ $8$ & $8$ $9$ $7$ $4$\\
Lookup & $010$ $t3$ $t4$ $t2$ $t6$ $t1$ $.$ &  $010$ $011$ $010$ $011$ $001$ $001$\\
Reverse Lookup & $t1$ $t6$ $t2$ $t4$ $t3$ $010$ $.$ & $010$ $011$ $010$ $011$ $001$ $001$\\
ReCopy & $4$ $7$ $9$ $8$ & $4$  $4$  $4$ $7$  $7$  $7$ $7$  $7$  $9$  $9$  $9$ $9$  $9$  $8$  $8$ $8$  $8$ $8$\\
Reverse ReCopy & $4$ $7$ $9$ $8$ & $8$ $8$  $8$ $8$  $8$ $9$  $9$  $9$ $9$  $9$  $7$  $7$  $7$ $7$  $7$ $4$  $4$  $4$\\
Inv ReCopy & $4$  $4$  $4$ $7$  $7$  $7$ $7$  $7$  $9$  $9$  $9$ $9$  $9$  $8$  $8$ $8$  $8$ $8$ & $4$ $7$ $9$ $8$\\
Inv Reverse ReCopy & $8$ $8$  $8$ $8$  $8$ $9$  $9$  $9$ $9$  $9$  $7$  $7$  $7$ $7$  $7$ $4$  $4$  $4$ & $4$ $7$ $9$ $8$\\
SCAN & look and run right & I\_LOOK I\_TURN\_RIGHT I\_RUN\\
\hline 
\end{tabular}
\caption{Input-output examples for each task (except CFQ).}
\label{table:task-examples}
\vspace{-4mm}
\end{table*}

To achieve the above desideratum 
and evaluate length generalization capability of different interlayer attention mechanisms, we set up ten synthetic probing task (see Table \ref{table:task-examples} and $\S$\ref{diagnostics}). Following prior work \cite{graves2014ntm, dehghani2018universal, Liang2021OutofDistributionGW}, we first consider the task of simply copying source texts in both forward and backward (reverse) directions. Following \citet{dubois-etal-2020-location}, we also consider compositional lookup table task \cite{liska2018memorize} in both directions. However, as we will show in $\S$\ref{diagnostics}, in these tasks the ideal attention position can be trivially determined from the decoding timestep alone---a condition (let us call it C1) that simply allows the relative positional attention \cite{shaw-etal-2018-self, dai-etal-2019-transformer} to perform perfectly given the right implementation. Thus, we propose new probing tasks involving repeated copy (ReCopy) and its variants to create settings where C1 is not satisfied. While there are already more synthetic tasks where C1 is not satisfied, our proposed tasks (ReCopy and its variants) are intended to be \textit{small} extensions over simple copy tasks such that the exact cause of model failure can be clearer compared to more complex tasks. Not only do we propose new probing tasks, but we also propose new strategies to tackle them. Prior models \cite{dehghani2018universal, dubois-etal-2020-location} already struggled in reverse copy or reverse lookup tasks. We introduce a technique of interpolating forward and reversed encoded representations to handle reverse direction even with simple relative attention (the technique is universally applicable to any seq2seq architecture). Moreover, we also propose new attention models, OneStep attention and monotonic location attention (our full model), to handle the proposed probing tasks on which the prior models fail. 
We also show that our models maintain comparable performance in the SCAN \cite{lake2018generalization} (a dataset for translating simple commands into sequences of actions) and CFQ \cite{keysers2020measuring} length splits (a dataset for query-to-SQL translation). 

\section{Probing Tasks}
\label{diagnostics}

We now describe the ten\footnote{Twelve including tasks in Appendix \ref{newtasks}.} probing tasks used in this paper. We present examples for each task in Table \ref{table:task-examples}. 

\textbf{Copy:} The copy task requires copying input tokens into the output tokens. In this case, the encoder-decoder network has to simply learn an identity function $(x=f(x))$. For this task we use a vocabulary of natural number tokens from $0$-$9$ (see Table \ref{table:task-examples} for an example). We generated $10,000$ training samples with sequence length in the range $5$-$10$. For the development set, we generated $2,000$ samples of sequence length $10$-$15$. For test sets, we generated a split with sequence length $15$, another split with sequence length $30$, and another with sequence length $100$. Each test split has $2,000$ samples.

\textbf{Reverse Copy:} In the reverse copy task, the model has to copy the input as above but in the reverse direction (see Table \ref{table:task-examples} for an example). This task is generated with the same parameters as the copy task.

\textbf{Lookup:} Lookup represents the ``Long Lookup Tables" task \cite{liska2018memorize} as made available in the code.\footnote{\url{https://github.com/i-machine-think/machine-tasks}\label{machine-task}} For any input like ``$001$ $t1$ $t2$ $t3$ .", the output for this task will be ``$v1$ $v2$ $v3$ $v4$" where $v1=001$, $v2=t1(001)$, $v3=t2(t1(001))$, and $v4=t3(t2(t1(001)))$. Here, $t1$, $t2$, and $t3$ are functions, each corresponding to some lookup table such that $t_i: \{0,1\}^3 \rightarrow \{0,1\}^3$ (for any natural number $i$). The task is generated using the supplied code.\footref{machine-task} The code generates a training split of approximately $9,000$ samples of lengths $\leq 6$. We consider three generated test splits that are of sequence lengths $7$, $9$, and $11$. The first test split has approximately $4,500$ samples whereas the others have approximately $5,000$ samples. The development split consists of about $500$ samples of sequence length $\leq 6$ and approximately $500$ samples of length $7$.

\textbf{Reverse Lookup:} Reverse Lookup represents the ``Long Lookup Tables Reverse" task \cite{liska2018memorize} as can be generated from the code.\footref{machine-task} For any input like ``$t1$ $t2$ $t3$ $001$ .", the output for this task will be ``$v1$ $v2$ $v3$ $v4$" where $v4=001$, $v3=t3(001)$, $v2=t2(t3(001))$, and $v1=t1(t2(t3(001)))$. Here, $t1$, $t2$, and $t3$ are lookup functions as before. The splits of this task are created similarly to those of the Lookup task described above.

\textbf{ReCopy:}\label{recopy} There is one thing that is common in the above tasks. For the forward tasks (Lookup, Copy), assuming that the encoder can keep the information of position $i$ at position $i$ after encoding with necessary contextualization (e.g., composition of previous functions in case of Lookup), the ideal encoding position to attend during decoding will always remain at the same constant distance from the decoding timestep. This is also true for the reversed tasks (Reverse Copy, Reverse Lookup) if the encoding is reversed. For example, to copy ``$4$ $7$ $9$ $8$", at timestep $1$ the model has to attend position $1$ to print $4$. Next, at timestep $2$ the model has to attend position $2$ to print $7$.  Thus, more generally, for any timestep $t$ the model has to attend an encoding position $i$ such that $i-t = c$ (where $c$ is some constant. In this example, $c = 0$). Even more generally, in all these tasks, the ideal position to attend can be determined just based on the decoding timestep $t$. For instance, for the above tasks, the ideal position of attention $i$ can be defined as a function over timestep as $i = f(t) = t+c$.  However, such a happy situation will not be maintained in more complex tasks. 

Thus, we decide to create a new set of diagnostic/probing tasks that are close to the previous tasks but precludes the possibility of determining the position of attention just from the timestep. With this motivation, first, we introduce the task ReCopy (Repeated Copy). In this task, the vocabulary includes natural numbers in the range $0$-$9$. If the input for the task is ``$4$ $7$ $9$ $8$", then the corresponding output will be ``$4$  $4$  $4$ $7$  $7$  $7$ $7$  $7$  $9$  $9$  $9$ $9$  $9$  $8$  $8$ $8$  $8$ $8$". Effectively, in this task, the model has to learn to not just copy but also to repeat the copied item for a certain frequency before copying the next item. There is a specific set of rules behind how many times an item should be repeated. That is, if the item is a number $\leq 3$ the model should print it once, if the item is a number $x$ such that $3 < x \leq 6$ the model should print it three times, and for any other number $> 6$, the model should print it five times. The splits and sample sizes for this task are similar to those of the copy task. 

Our intention here is to make a small extension of the copy task that avoids determination of the attention position purely from the timestep but without introducing any additional sources of difficulty so that the causes of failures can be disentangled more easily. For instance, if a model succeeds in the copy task but fail in ReCopy we can now reasonably infer that its cause of failure is the specific difficulty introduced in ReCopy. Note that if we made ReCopy a bit simpler by requiring each number to be copied and repeated for a uniform frequency, then the determination of the ideal position for attention will again become trivially possible just from a decoding timestep; thus ReCopy requires repeating with varying frequency depending on which number is being copied. 

\textbf{Reverse ReCopy:} The Reverse ReCopy task is similar to the ReCopy task in all aspects other than the fact that the copying takes place in the reversed direction (see example in Table \ref{table:task-examples}). The task splits are generated in the same way as in the Copy task. 

\textbf{Inv ReCopy:} The Inv ReCopy task (Inverse ReCopy)  is similar to the ReCopy task in all aspects other than the fact that the inputs and outputs are inverted (see example in Table \ref{table:task-examples}). The task splits are generated in the same way as in the Copy task. 

\textbf{Inv Reverse ReCopy:} The Inv Reverse ReCopy task (Inverse Reverse ReCopy)  is similar to the Reverse ReCopy task in all aspects other than the fact that the inputs and outputs are inverted (see example in Table \ref{table:task-examples}). The task splits are generated in the same way as in the Copy task. 

\textbf{SCAN:} SCAN is a popular dataset used for testing systematic generalization \cite{lake2018generalization}. It involves the task for translating simple commands into a sequence of actions. We explore its original length generalization split.

\textbf{CFQ:} CFQ is a realistic semantic parsing dataset \cite{keysers2020measuring} proposed for evaluating compositional generalization. We explore its length generalization split.

We also propose and explore two additional probing tasks (\textbf{DeDupe} and \textbf{PosRetrieve}) in Appendix \ref{newtasks}. 

\section{Seq2Seq General Framework}
\label{Methods}
A seq2seq model can be formalized as a function $F_{seq2seq}: \mathbb{N}^{s} \rightarrow \mathbb{N}^{z}$ that maps some input sequence $x_{1:s} = (x_1, x_2, \dots, x_s)$ of length $s$ to an output sequence $y_{1:z} = (y_1, y_2, \dots, y_z)$ of length $z$. Here each element in $x_{1:s}$ and $y_{1:z}$ is a natural number that indexes some distinct token from a vocabulary. $F_{seq2seq}$ constitutes two major components: an encoder function ($F_{enc}$) and a decoder function ($F_{dec}$). The encoder $F_{enc}: \mathbb{N}^{s} \rightarrow {\rm I\!R}^{s \times d}$ maps the initial token indices $x_{1:s}$ to a sequence of hidden state representations $e_{1:s} = (e_1,e_2,\dots,e_s)$ (where any $e_i \in {\rm I\!R}^d$). The decoder $F_{dec}: \mathbb{N}^* \times \mathbb{N} \rightarrow \mathbb{N}$ generates the output sequence $y_{1:z}$ recursively one token at a time, typically in an autoregressive manner. That is, at
each timestep $t$ (beginning at $t=1$), $F_{dec}$ takes as input the history of all previously generated tokens $H^{t-1} = (go, y_1, y_2, \dots, y_{t-1})$ and the last generated token index $y_t$ and outputs $y_{t+1}$. $H^0$ is initialized with $(go)$ where $go$ represents the index of a special token that marks the beginning of the generation.  

One salient component within the decoder is an interlayer (cross) attention function \cite{bahdanau2015neural} that allows the decoder to interact and retrieve encoded state information. The decoder, in timestep $t$, will typically create history-contextualized representation $h^{t-1} \in {\rm I\!R}^d$ (compressing $H^{t-1}$). Let query $q_{t-1} = f_q(h_{t-1})$, keys $k_i = f_k(e_{i})$, and values $v_i = f_v(e_{i})$ ($\forall i \in \{1,\dots,s\})$ where $f_q, f_k,$ and $f_v$ are linear transformations ($f_{q|k|v}: {\rm I\!R}^d \rightarrow {\rm I\!R}^d$).
In the attention layer, the query interacts with the keys to score each corresponding value. A weighted sum of values based on those scores is then computed as the result of the attention function. This allows the decoder to dynamically and softly retrieve information from any position in the encoded representations $e_{1:s}$ at any timestep. For our work, we explore a variety of cross-attention functions which we discuss below.

\section{Prior Approaches to Cross-Attention}
\label{prior}
\subsection{Content Attention}
\label{content_attention}
As a baseline, we use the popular scaled inner dot-product query-key based attention as used in \citet{vaswani2017attention}:
\begin{equation}
    c_{ti} = \frac{<q_t , k_i>}{\sqrt{d}}, 
    \label{vanilla_dot}
\end{equation}
\begin{equation}
    a_{ti} = \frac{\exp(c_{ti})}{\sum_{j=1}^s \exp(c_{tj})},\;  o_{t} = f_o(\sum_{j=1}^s a_{tj} \cdot v_j),
\end{equation}
where $f_{o}: {\rm I\!R}^d \rightarrow {\rm I\!R}^d$ is a linear transformation, $c_{ti}, a_{ti} \in {\rm I\!R}$ and $o_t \in {\rm I\!R}^d$. Note that this is a fully \textit{content-based attention} because it does not explicitly use any position or distance-related information about the query or keys.

\subsection{Relative Attention}
\label{relative_attn}
As another baseline, we use the relative attention mechanism\footnote{Initially the idea was introduced in \citet{shaw-etal-2018-self}.} as used in \citet{dai-etal-2019-transformer}. Effectively, a sinusoidal positional encoding \cite{vaswani2017attention, dai-etal-2019-transformer} is first used to create embeddings for each relative distance. Let $pe_{k} \in {\rm I\!R}^d$ represent the embedding for the distance $k \in \mathbb{Z}$. Then, the relative position attention creates a query-key score sequence based on the corresponding relative distances between the query and the keys:
\begin{equation}
    r_{ti} = \frac{<(q_t + b_2), pe_{i-t}>}{\sqrt{d}}
\end{equation}
where $b_2 \in {\rm I\!R}^d$ is a bias for position attention and $r_{ti} \in {\rm I\!R}$. This is integrated with content-based attention by modifying Eqn. \ref{vanilla_dot} in $\S$\ref{content_attention} as:
\begin{equation}
    c_{ti} = \frac{<(q_t + b_1), k_i>}{\sqrt{d}} + r_{ti}
\end{equation}
$b_1 \in {\rm I\!R}^d$ is a bias for content-based attention. Everything else is kept the same as was for content-based attention.

\subsection{Location Attention}
\label{locattn}
Location attention, as introduced in \citet{dubois-etal-2020-location}, is primarily a form of attention based only on the positions of the encodings $e_{1:s}$; however, it is more expressive than the relative positional attention. Here we discuss the details of location attention with some refinements. \citet{dubois-etal-2020-location} introduced a method to determine the locational ``center of focus" for attention which is made to resemble human attention in visual search in how even when it focuses on a specific part of the input, it also perceives neighboring parts due to the eccentricity effect \cite{Carrasco1995}. 
Let $\mu_t \in {\rm I\!R}$ represent the center of focus such that positions close to $\mu_t$ get higher attention than those farther away. With such $\mu_t$, an attention spread can be modeled by using $\mu_t$ as a mean in a Gaussian distribution:
\begin{equation}
    \lambda_{ti} = \exp\left(-\frac{(i-\mu_t)^2}{2 \cdot \sigma_t^2}\right)
\end{equation}
Here $\sigma_t$ is the standard deviation, which determines the spread of the attention focus. However, using raw values of $i$ and $\mu_t$ is not ideal because the range of values (especially of $i$) can differ based on sequence length. This becomes more problematic for unseen length generalization. 
Thus, the formalism is modified as follows:
\begin{equation}
    \lambda_{ti} = \exp\left(-\frac{(norm(i)-clamp(\mu_t))^2}{2 \cdot \sigma_t^2}\right)
\end{equation}
where:
\begin{equation}
    norm(i) = \frac{i-1}{max(1,s-1)}
\end{equation}
\begin{equation}
    clamp(\mu_t) = max(0 + m \cdot \mu_t, min(1 + m \cdot \mu_t, \mu_t))
\end{equation}
Note that the encoder position index ranges from $1$ to $s$ where $s$ is the sequence length. The $norm()$ function squeezes any position index $i$ into the range $[0,1]$ no matter the sequence length. Further, the $clamp()$ function enforces $\mu_t$ to be approximately within $[0,1]$ which is the possible range of positions that can be attended. Following \citet{dubois-etal-2020-location}, $m$ in $clamp()$ acts as a negative slope ($m=0.01$) to add some ``leakiness" similar to LeakyReLU. Note that the result is a PDF over the whole real number set whereas only 
the discrete positions of the encoding matter. Thus, $\lambda_{ti}$ can be further normalized  to get a discretized probability measure over only the relevant positions:
\begin{equation}
    \lambda'_{ti} = \frac{\lambda_{ti}}{\sum_{j=1}^s \lambda_{tj}}
\end{equation}
This gives the location attention. Below we discuss how to obtain $\mu_t$ and $\sigma_t$. First, a transformation over the decoder hidden state $h_t$ is created as $l_t = f_l(h_t)$ where $f_l: {\rm I\!R}^d \rightarrow {\rm I\!R}^d$ is a linear transformation.\footnote{\citet{dubois-etal-2020-location} used a GRU for $f_l$. However, in our experiments we removed it because we did not find it effective.} Next, $\sigma_t$ is computed as:
\begin{equation}
    \sigma_t = \frac{ReLU(f_{\sigma_t}(l_t)) + min_{\sigma_t}}{s}
\end{equation}
Here $f_{\sigma_t}: {\rm I\!R}^d \rightarrow {\rm I\!R}$ is a linear transform and $min_{\sigma_t}$ is the minimum value for $\sigma_t$. Next, $\mu_t$ is computed by taking some steps (in either forward or backward direction) with respect to some reference position ref$_{t}$. Given that the $norm(.)$ function will squeeze any discrete position index into a continuous number in $[0,1]$, ref$_{t}$ can also be treated to be in $[0,1]$. Formally, $\mu_t$ is computed as: 
\begin{equation}
    \mu_t = \text{ref}_{t} + \text{stepsize} \cdot \text{steps}_t
\end{equation}
Here, stepsize is $\frac{1}{max(1,s-1)}$. For the reference point ref$_{t}$, different possible choices can be considered. One possible choice is the previous attended position $pa_{t-1}$ which is computed as $pa_{t-1} = \sum_{i=1}^s \alpha_{t-1i} \cdot norm(i)$ where $\alpha_{t-1i}$ represents the interlayer attention at the previous timestep ($t-1$) to the encoding position $i$. Essentially, with this setup, the attention model can move left or right with respect to previously attended position. Another choice for the reference point is to simply make a neural network-based logistic prediction to choose any arbitrary position $b_t$ in $[0,1]$ as $b_{t} = sigmoid(f_b(l_t))$ where $f_{b}: {\rm I\!R}^d \rightarrow {\rm I\!R}$ is a linear transform. Here $b_{t}$ can also help initialize ref$_t$ to adaptively learn to start attending from the beginning of the encoding ($i=1$) or the end of the encoding ($i=s$) (or even somewhere in-between if needed) based on the task. Ultimately, we can allow the model itself to learn to choose or combine both $pa_{t-1}$ and $b_t$ as needed:
\begin{equation}
    \text{ref}_{t} = g_t \cdot pa_{t-1} + b_t 
    \label{ref_comp}
\end{equation}
with $g_t$ being a real scalar in $[0,1]$ functioning as a gate that decides to keep or ignore $pa_{t-1}$. It is computed as $g_t = sigmoid(f_g(l_t))$ where $f_{g}: {\rm I\!R}^d \rightarrow {\rm I\!R}$ is a linear transform. Next the steps to take (i.e., steps$_t$) with respect to the reference point are determined as follows:
\begin{equation}
    \text{steps}_t = \text{softstair}(f_{step}(l_t))
    \label{lab:step}
\end{equation}
where $f_{step}: {\rm I\!R}^d \rightarrow {\rm I\!R}$ is again a linear transform and softstair is an activation function that pushes the output towards an integer:
\begin{equation}
   \text{softstair}(x) = \lfloor x\rfloor + sigmoid(\tau \cdot (x-\lfloor x\rfloor-0.5))
\end{equation}
$\tau$ is a temperature hyperparameter which is set to $20$ like in \citet{dubois-etal-2020-location}. Last, the attention is computed as a convex combination of content attention and location attention:
\begin{equation}
    a_{ti} = mix_{ti} \cdot \left(\frac{\exp(c_{ti})}{\sum_{j=1}^s \exp(c_{tj})}\right) + (1-mix_{ti}) \cdot \lambda'_{ti}
    \label{mixup}
\end{equation}
\begin{equation}
    mix_{ti} = sigmoid(\beta f_{mix} (h_t))
\end{equation}
Here $f_{mix}: {\rm I\!R}^d \rightarrow {\rm I\!R}$ is a linear transform and $c_{ti}$ corresponds to the content-based pre-normalized attention scores as computed in Eqn. \ref{vanilla_dot}. In some cases, we might want to ignore the content attention focusing purely on location attention. In such cases, we can set $a_{ti} = \lambda'_{ti}$.


\section{Proposed Approaches to Cross-Attention}
In this section, we first present the limitations of the prior approaches discussed above and then present (in a {\em bottom-up} manner) our proposed changes that address them. 
\subsection{Limitations of Prior Approaches}
\label{limits}

\textbf{Limitation 1 (handling reverse tasks):} As noted earlier (see ReCopy task description in $\S$\ref{diagnostics}), in some tasks like Copy or Lookup, the target cross-attention position is always at the same constant relative distance from the timestep. In such cases, the inductive bias from the relative attention ($\S$\ref{relative_attn}) can be especially fruitful. However, this setup is not maintained by default (without reversing the encoding or the input in the model), in the reverse directions of the tasks (Reverse Copy or Reverse Lookup). Consider transforming ``$4$ $7$ $9$ $8$" to ``$8$ $9$ $7$ $4$". In this case to print $8$ in timestep $t=1$, the model needs to attend to encoding position $i=4$. Thus, the relative distance will be $i-t=3$. However, for printing $9$ in timestep $t=2$, the model needs to attend to the encoding position $i=3$. Then the relative distance will be $i-t=1$. Thus, the ideal relative distance can vary with timestep and also depends on the source sequence length. These facts make it a struggle for relative attention, by default, to work on reverse tasks. In theory, location attention is equipped to handle reverse tasks - it has to just initialize $b_t$ as $1$ and $g_t$ as $0$ when $t=1$. This will set $ref_t=1$, i.e., the reference position will be the end of the input sequence. From that point location attention can learn to take steps backward one by one using previous attention ($pa_{t-1}$) as the reference position if needed. However, in practice, location attention still tends to be empirically brittle and have been shown to fail the reverse lookup task \cite{dubois-etal-2020-location}.   

\textbf{Limitation 2 (handling ReCopy and beyond):} As discussed in  $\S$\ref{diagnostics} (see ReCopy description), tasks like ReCopy, Reverse ReCopy, or their inverted variants are specifically designed to serve as settings in which the ideal attention position can vary from timestep to timestep (no matter if the encoding positions are reversed or not). Thus, this setting becomes hard for relative attention. Location attention, again, can theoretically address these situations given its flexibility to keep track of past attended position and ability to take any arbitrary steps in reference to past attended position dependent on the decoder state. Nevertheless, as mentioned earlier, in practice location attention turns out to be relatively brittle. Moreover, its use of soft sigmoid-based gating for making decisions at different stages of the model can lead to higher error accumulation and lower robustness to increasing lengths of data.

\subsection{Bidirectional Relative Attention}
\label{dir_relative_attn}
First, we propose a simple approach to extend relative attention in a manner that addresses limitation 1. We note that if the task is, e.g., reverse copy, we can simply reverse the encoded sequence representations after encoding. Once done so, from the perspective of the decoder, the task becomes equivalent to forward copy. Nevertheless, in practice, we will not know ahead of time whether we are facing a task where forward version of the encoding is more ideal or the reversed version of the encoding. Thus, we use a gating mechanism that interpolates (make a convex combination of) the two directions of encoding so that the model can adaptively decide whether to reverse the encodings or not: 
\begin{equation}
    e^{rev}_{1:s} = \text{reverse}(e_{1:s}),\; \alpha_{dir} = \text{sigmoid}(\beta \cdot f_{dir}(e_{cls}))
\end{equation}
\begin{equation}
    \forall i \in \{1,\dots,s\}  \mbox{   }\mbox{   }\mbox{   } e^{dir}_i = \alpha_{dir} \cdot e_i + (1-\alpha_{dir}) \cdot e^{rev}_i
\end{equation}
$\beta$ is a scalar (acting as a temperature), $f_{dir}: {\rm I\!R}^d \rightarrow {\rm I\!R}^1$ is a linear layer, and $e_{cls} \in {\rm I\!R}^d$ is a vector representation of the whole sequence $e_{1:s}$ - it can be implemented in multiple ways (we explain our implementation in Appendix \ref{exp_details}). After this we use the same strategy as in $\S$\ref{relative_attn} but using key and value transformations over $e^{dir}$ instead of $e$. This trick can also be useful in more practical tasks like translation where the source and target language may have different reading orders.  Note that $e^{dir}$ is different from outputs from models like bidirectional RNNs. Unlike here, in a bidirectional RNN, the encoded tokens remain in the same positions; only the contextualizing information comes from different directions. Also, note that this strategy is as general purpose as introducing bidrectionality to RNNs. Moreover, we are allowing neural networks to dynamically predict the direction for a given input through the gating mechanism; thus, avoiding infusion of task-specific knowledge of ideal direction of attention.\footnote{While this strategy may appear obvious, it is still not explored so far to our knowledge. Moreover, theoretical motivation does not always translate well to empirical performance. For instance, Location Attention struggles in reverse tasks despite having the theoretical capacity for reverse attention as discussed in $\S$\ref{limits}. So empirical benefit of this strategy is not a priori obvious and deserves the investigation that we do here.}  

\subsection{OneStep Attention}
\label{onestep}
As discussed in $\S$\ref{limits}, fixing limitation 1 by reversing the encodings (as in $\S$\ref{dir_relative_attn}) still does not address limitation 2. Concerned with limitation 2, we move away from simple relative positional attention and instead seek to make adjustments over location attention to address its potential issues (see $\S$\ref{limits}). As a result we propose a new attention model - OneStep attention. Below we enumerate the main adjustments over location attention (from $\S$\ref{locattn}): 
\begin{enumerate}
\vspace{-2mm}
    \item OneStep attends to key-value transformations of $e^{dir}_{1:s}$ instead of $e_{1:s}$ similar to $\S$\ref{dir_relative_attn}. 
    \vspace{-2mm}
    \item The computation of ref$_t$ is simplified as: $\text{ref}_t = pa_{t-1}$
    \vspace{-2mm}
    \item The activation function in Eqn. \ref{lab:step} to sigmoid from softstair: $\text{steps}_t = \text{sigmoid}(f_{step}(l_t))$
\end{enumerate}
\vspace{-1mm}
\textbf{First Change:} The first change follows from $\S$\ref{dir_relative_attn} and is motivated to address limitation 1. 

\textbf{Second Change:} The second change is motivated by a number of factors. First, due to the incorporation of the first change, the role of $b_t$ from Eqn. \ref{ref_comp} is severely undermined. It is not anymore necessary for $b_t$ to initialize the starting position of attention to handle reverse tasks. Besides that the usefulness of $b_t$ can be limited.\footnote{It can be still useful in special cases when the model has to attend some $x$ position from the end in one timestep and some $y$ position from the beginning in another.} It is motivated for percentile attention in \citet{dubois-etal-2020-location} which may not be as relevant or can be accomodated by content attention mixing (Eqn. \ref{mixup}). So we removed it. To reduce error-accumulation we also remove the gating $g_t$ over $pa_{t-1}$; thus ultimately setting $\text{ref}_t = pa_{t-1}$. It removes the models capacity for attending to some specific absolute position from the beginning/end but this capacity is also lacking from relative attention and is not currently required by most of our tasks. We keep investigation to incorporating this capacity better in the future. Currently, absolute positional encoding in the encoder combined with content attention mixing can still accommodate for the lack to an extent.

\textbf{Third Change:} In the third change, we replace softstair with a sigmoid for the step computation. The sigmoid function enforces the model to softly choose between either taking a single step forward (steps$_t = 1$) or none (steps$_t = 0$). We added this change because giving unrestricted freedom in determining the steps can make it harder for the model to learn the right function. Particularly in most of our current diagnostic tasks, it is sufficient to learn to make bounded steps in $[0,1]$ with respect to the past attended position.  While this choice is perhaps not ultimately ideal, it helps us evaluate the breaking points of the Location Attention framework better. Regardless, even after this restriction, OneStep can be still powerful enough to simulate a windowed version of relative attention (if it takes a single step in every timestep) \cite{shaw-etal-2018-self}. Moreover, a sufficiently powerful encoded representation can, in theory, always reorganize or permute the input information to accommodate for this restriction. Besides, content attention mixing (Eqn. \ref{mixup}) can break the monotonicity of OneStep\footnote{By itself, without content attention mixing, OneStep is monotonic because in it, the center of focus can only move forward with time.} and make it more flexible. 

\subsection{Monotonic Attention}
\label{monoattn}
In some tasks, it can be easier to learn to take bigger steps at the level of interlayer attention instead of expecting the encoder to permute the source input appropriately. So, we create another attention function where we relax the constraints in OneStep by changing the steps computation as:  
\begin{equation}
    \text{steps}_t = g \cdot \text{sigmoid}(f_{step}(l_t)) + (1-g) \cdot ReLU(f_{step}(l_t))
    \label{sigmoid-gate}
\end{equation}
Here, $g = \text{sigmoid}(p)$ where $p \in {\rm I\!R}$ is a model parameter.\footnote{In future, it can be better to have $g$ dependent on the input encoding such as $e^{cls}$ in case we want a multi-tasking model.} As we can see, with this setup we can allow the model itself to learn to prefer either taking controlled steps with a sigmoid or possibly bigger steps with a ReLU. We still use ReLU activation to keep the attention monotonic (i.e., the attention mechanism can only make forward steps) similar to OneStep for reasons discussed in $\S$$\S$\ref{onestep} (in Third Change).

 \begin{table*}[t]
\small
\centering
\def\arraystretch{1.2}
\begin{tabular}{  l | l l l | l l l | l l l | l l l} 
\hline
\textbf{Model} & \multicolumn{3}{c}{\textbf{Copy}} & \multicolumn{3}{|c}{\textbf{Reverse Copy}} & \multicolumn{3}{|c}{\textbf{Lookup}} & \multicolumn{3}{|c}{\textbf{Reverse Lookup}}\\
(Length Splits)  & $15$ & $30$ & $100$ & $15$ & $30$ & $100$ & $7$ & $9$ & $11$ & $7$ & $9$ & $11$\\
\hline
Content & $0$ & $0$ & $0$ & $0$ & $0$ & $0$ & $33.3$ & $0$ & $0$ & $3.7$ & $0$ & $0$\\
Relative & $\mathbf{100}$ & $\mathbf{100}$ & $\mathbf{100}$ & $0$ & $0$ & $0$ & $\mathbf{100}$ & $\mathbf{100}$ & $\mathbf{100}$ & $78.2$ & $0.8$ & $0.4$\\
LocAttn &  $99.8$ & $0$ & $0$ & $0.7$ & $0$ & $0$ & $\mathbf{100}$ & $9.4$ & $0$ & $13.3$ & $0$ & $0$\\
\hline
Ours \\
\hline
Bi-Relative & $\mathbf{100}$ & $\mathbf{100}$ & $\mathbf{100}$ & $\mathbf{100}$ & $\mathbf{100}$ & $\mathbf{100}$ & $\mathbf{100}$ & $\mathbf{100}$ & $\mathbf{100}$ & $\mathbf{100}$ & $\mathbf{100}$ & $\mathbf{100}$\\
OneStepAttn & $\mathbf{100}$ & $\mathbf{100}$ & $\mathbf{100}$ & $\mathbf{100}$ & $\mathbf{100}$ & $\mathbf{100}$ & $\mathbf{100}$ & $\mathbf{100}$ & $\mathbf{100}$ & $\mathbf{100}$ & $\mathbf{100}$ & $\mathbf{100}$\\
MonoAttn & $\mathbf{100}$ & $\mathbf{100}$ & $\mathbf{100}$ & $\mathbf{100}$ & $\mathbf{100}$ & $\mathbf{100}$ & $\mathbf{100}$ & $98$ & $29.9$ &  $28.5$ & $0$ & $0$ \\
\hline 
\textbf{Model} & \multicolumn{3}{c}{\textbf{ReCopy}} & \multicolumn{3}{|c}{\textbf{Reverse ReCopy}} & \multicolumn{3}{|c}{\textbf{Inv ReCopy}} & \multicolumn{3}{|c}{\textbf{Inv Reverse ReCopy}}\\
(Length Splits)  & $15$ & $30$ & $100$ & $15$ & $30$ & $100$ & $15$ & $30$ & $100$ & $15$ & $30$ & $100$\\
 \hline
Content & $19.1$ & $0$ & $0$ & $25$ & $0$ & $0$ & $0.05$ & $0$ & $0$ & $0$ & $0$ & $0$\\
Relative & $43.1$ & $0$ & $0$ & $0.1$ & $0$ & $0$ & $75.9$ & $0$ & $0$ & $0$ & $0$ & $0$\\
LocAttn & $79.6$ & $0$ & $0$ & $19.7$ & $0$ & $0$ & $99.4$ & $58.8$ & $0$ & $97.9$ & $0.3$ & $0$\\
\hline
Ours \\
\hline
Bi-Relative & $33.4$ & $0$ & $0$  & $35.3$ & $0$ & $0$ & $69.8$ & $0$ & $0$ &  $71.3$ & $0$ & $0$\\
OneStepAttn & $\mathbf{100}$ & $\mathbf{100}$ & $\mathbf{100}$ & $\mathbf{100}$ & $\mathbf{100}$ & $\mathbf{100}$ & $0.1$ & $0$ & $0$ & $0$ & $0$ & $0$\\
MonoAttn & $\mathbf{100}$ & $\mathbf{100}$ & $\mathbf{100}$ & $\mathbf{100}$ & $\mathbf{100}$ & $\mathbf{100}$ & $\mathbf{100}$ & $\mathbf{100}$ & $\mathbf{98.8}$ & $\mathbf{100}$ & $\mathbf{99.9}$ & $\mathbf{98.3}$\\
\hline 
\end{tabular}
\caption{Accuracy of the models on different length generalization splits in different algorithmic diagnostic / probing tasks. We present the median of five runs on different seeds. We bold the best results.}
\label{table:diagnostics}
\vspace{-2mm}
\end{table*}

\begin{table}[t]
\small
\centering
\def\arraystretch{1.2}
\begin{tabular}{  l | l | l} 
\hline
\textbf{Model} & \textbf{SCAN (Len.)} & \textbf{CFQ (Len.)}\\
 \hline 
Content & $17.61 \pm 4.07$ & $\mathbf{62,14 \pm 0.88}$\\
Relative & $19.21 \pm 5.52$ & $56.64 \pm 1.84$\\
Mix LocAttn & $20.74 \pm 5.69$ & $44.83 \pm 9.45$\\
\hline
Ours\\
\hline
Bi-Relative & $8.41 \pm 1.21$ & $59.48 \pm 1.54$\\
Mix OneStepAttn & $\mathbf{29.51 \pm 9.46}$ & $60.65 \pm 3.74$\\
Mix MonoAttn &  $21.08\pm 7.17$ & $60.32 \pm 3.58$\\
\hline 
\end{tabular}
\caption{Accuracy on SCAN length split and CFQ length split. We report the mean and standard deviation of $5$ runs for SCAN and of $3$ runs for CFQ. We bold the best results.}
\label{table:scan}
\vspace{-5mm}
\end{table}

\section{Experimental Setup}
Similar to \citet{dubois-etal-2020-location}, we use a Bidirectional GRU \cite{chung2014gru} based seq2seq model as the base for all the attention mechanisms. We explain more architectural details and hyperparameters in the Appendix \ref{exp_details}. 

\textbf{Nomenclature: } In Tables \ref{table:diagnostics} and \ref{table:scan}, we use the term \textit{Content} to refer to content attention ($\S$\ref{content_attention}), \textit{Relative} to refer to relative attention ($\S$\ref{relative_attn}), and \textit{Bi-Relative} for bi-directional relative attention ($\S$\ref{dir_relative_attn}). We use the terms \textit{LocAttn}, \textit{OneStepAttn}, and \textit{MonoAttn} for location attention ($\S$\ref{locattn}), OneStep Attention ($\S$\ref{onestep}), and monotonic attention ($\S$\ref{monoattn})  respectively if they are used without mixing content attention (i.e., replacing Eqn. \ref{mixup} with $a_{ti} = \lambda'_{ti}$). Otherwise, we use the terms \textit{Mix LocAttn}, \textit{Mix OneStepAttn}, and \textit{Mix MonoAttn} when mixing with content attention is done (i.e., Eqn. \ref{mixup} is kept as described). We generally use the unmixed variants on the simpler diagnostic tasks (Lookup, Copy, or ReCopy-based tasks) because position-based attention is what is mainly relevant for the tasks.  

\textbf{Evaluation:} We calculate the sequence-level accuracy of our models. Any generated output gets a score of $1$ if and only if it matches exactly with the given target output. 

\textbf{On the EOS problem:} The EOS token is a special marker  that a model needs to generate to signify the end of sequence. In similar contexts, some prior works have tried to make the evaluation less stringent \cite{dubois-etal-2020-location, Newman2020TheED} by terminating the model generation based on the oracle EOS position or by truncating oracle sequence based on predicted EOS position. We do not modify the evaluation in any such non-standard manner. Generally, we do not find EOS prediction to be a problem. If the inductive bias is suitable for the task, our models learn to generalize near perfectly without us needing to incorporate any separate mechanism to predict EOS properly. 

 \begin{table*}[t]
\small
\centering
\def\arraystretch{1.2}
\begin{tabular}{  l | l l l | l l l | l l l | l l l} 
\hline
\textbf{Model} & \multicolumn{3}{c}{\textbf{Copy}} & \multicolumn{3}{|c}{\textbf{Reverse Copy}} & \multicolumn{3}{|c}{\textbf{Lookup}} & \multicolumn{3}{|c}{\textbf{Reverse Lookup}}\\
(Length Splits)  & $15$ & $30$ & $100$ & $15$ & $30$ & $100$ & $7$ & $9$ & $11$ & $7$ & $9$ & $11$\\
\midrule
OneStepAttn & $\mathbf{100}$ & $\mathbf{100}$ & $\mathbf{100}$ & $\mathbf{100}$ & $\mathbf{100}$ & $\mathbf{100}$ & $\mathbf{100}$ & $\mathbf{100}$ & $\mathbf{100}$ & $\mathbf{100}$ & $\mathbf{100}$ & $\mathbf{100}$\\
$-$Step $2$ & $\mathbf{100}$ & $1.4$ & $0$ & $\mathbf{100}$ & $98.6$ & $0$ & $99.2$ & $2.34$ & $0$ & $99.8$ & $0$ & $0$\\
$-$Step $3$ & $6.9$ & $0$ & $0$ & $0$ & $0$ & $0$ & $41.9$ & $0$ & $0$ & $22.9$ & $0$ & $0$\\
$-$Sigmoid & $\mathbf{100}$ & $\mathbf{100}$ & $99.8$ & $\mathbf{100}$ & $\mathbf{100}$ & $\mathbf{100}$  & $\mathbf{100}$ & $74.3$ & $0.3$ & $19.1$ & $0$ & $0$\\
\midrule
\textbf{Model} & \multicolumn{3}{c}{\textbf{ReCopy}} & \multicolumn{3}{|c}{\textbf{Reverse ReCopy}} & \multicolumn{3}{|c}{\textbf{Inv ReCopy}} & \multicolumn{3}{|c}{\textbf{Inv Reverse ReCopy}}\\
(Length Splits)  & $15$ & $30$ & $100$ & $15$ & $30$ & $100$ & $15$ & $30$ & $100$ & $15$ & $30$ & $100$\\
 \hline
OneStepAttn & $\mathbf{100}$ & $\mathbf{100}$ & $\mathbf{100}$ & $\mathbf{100}$ & $\mathbf{100}$ & $\mathbf{100}$ & $0.1$ & $0$ & $0$ & $0$ & $0$ & $0$\\
$-$Step 2 & $15.9$ & $0$ & $0$ & $16.9$ & $0$ & $0$ & $\mathbf{95.5}$ & $0$ & $0$ & $\mathbf{96.2}$ & $0$ & $0$\\
$-$Step 3 & $\mathbf{100}$ & $\mathbf{100}$ & $\mathbf{100}$ & $\mathbf{100}$ & $\mathbf{100}$ & $99.9$ & $40.3$ & $0$ & $0$ & $45$ & $0$ & $0$\\
$-$Sigmoid & $\mathbf{100}$ & $\mathbf{100}$ & $\mathbf{100}$ & $\mathbf{100}$ & $\mathbf{100}$ & $\mathbf{100}$ & $0$ & $0$ & $0$ & $22$ & $0$ & $0$\\
\hline 
\end{tabular}
\caption{Accuracy of ablations over OneStepAttn in different length generalization splits in different algorithmic diagnostic/probing tasks. We present the median of five runs on different seeds. We bold the best results.}
\label{table:critical_ablation}
\vspace{-4mm}
\end{table*}

\section{Experimental Results}

In Table \ref{table:diagnostics} we show the results of our different attention strategies on all our diagnostic tasks except SCAN and CFQ. The results are close to what we would expect a priori. Pure content attention (Content) without more explicit guidance from any positional information suffers in all the tasks. Relative attention (Relative) does well in the forward copy and lookup tasks but it fails in the reversed tasks for the reasons discussed in $\S$\ref{dir_relative_attn}. It also fails in the ReCopy-based tasks. This is consistent with our discussed limitations of prior works in $\S$\ref{limits}. Also, consistent with this discussion, we find our implementation of location attention (LocAttn) to struggle in all the tasks. 

Bidirectional relative attention (Bi-Relative) succeeds on both forward and reverse directions of copy and lookup tasks. This is aligned with our motivation for designing it ($\S$\ref{dir_relative_attn}). However, Bidirectional relative attention still does not alleviate the second limitation ($\S$\ref{limits}) and thus,  fail in the ReCopy-based tasks. 

OneStep attention (OneStepAttn) succeeds nearly on all tasks except the inverted variations of the ReCopy tasks. The Copy tasks and Lookup tasks are easy to learn for OneStep attention because in either tasks it has to simply learn to take one step forward relative to the past attended position in every timestep. The ReCopy and Reverse ReCopy is slightly more complicated but still not too hard to learn. In these cases, the model has to learn to wait (predict $steps_t=0$) while the decoder is repeating previous generations. The attention model has to then predict $steps_t = 1$ to move one step forward in the encoding positions after the repetition of the content from the past attended position is complete. Thus, the OneStep strategy is suitable for the ReCopy and Reverse ReCopy tasks as well. 

However, the OneStep strategy faces an issue for the inverted versions of the tasks. Consider an Inv ReCopy sample where the input is  “4 4 4 7 7 7 7 7 9 9 9 9 9 8 8 8 8 8” and the output is  “4 7 9 8”. In this case, one way to solve this would be for the encoder to radically re-organize the positions of the input information. But if the encoder fails to do that and keeps the encoded information close to its original position, OneStep attention, by itself, is ill-equipped for the task. In the given example, after printing $4$ from encoding position $1$, in the next timestep it has to take not just one but three steps forward. OneStep attention cannot do that because its number of steps is constrained by a sigmoid. 

In contrast to OneStep attention, monotonic attention (MonoAttn) is more flexible allowing bigger steps when needed. As such, monotonic attention is able to solve Inv ReCopy tasks that OneStep could not. It also performs perfectly on copy tasks and ReCopy tasks in both directions. However, it fails on the lookup tasks. It seems that its increased flexibility (loosened inductive bias) and its possibility to make more uncontrolled steps (which are unnecessary for the lookup tasks) also at the same time make it more confused when trying to learn the lookup tasks in a length-generalizing manner. 

Ultimately, both OneStep attention and monotonic attention perform better than any of the other attention models. Both solves $6$ out of the $8$ tasks in Table \ref{table:diagnostics} with $100\%$ accuracy. However, we also discover a trade-off - the restricted steps of OneStep attention preclude it from solving the inverted versions of ReCopy tasks whereas the more unconstrained steps of monotonic attention manages the inverted ReCopy tasks but at the cost of the lookup tasks. 

In Table \ref{table:scan}, we present the results on SCAN. We find location attention and our extensions of it (OneStep attention or monotonic attention) to generally also perform better on the task of translating simple commands into sequences of actions than other forms of interlayer attention even though they are not designed explicitly keeping the structure of SCAN task in mind. OneStep attention (Mix OneStepAttn) does particularly better than the others in SCAN. In the same table, we also present the results on CFQ. Interestingly, the basic position-encoding-less version of inter-layer attention does the best here. However, both OneStep and monotonic attention keep up with it better than others - like location attention or unidirectional relative attention. 

\subsection{Additional Analyses}

\textbf{Ablations:} In Table \ref{table:critical_ablation}, we show some of the main ablations of OneStep Attention. $-$Step $2$ represents using the more sophisticated location attention variant of ref$_t$ computation (Eqn. \ref{ref_comp}) instead of the proposed $ref_t = pa_{t-1}$ change in step $2$ in $\S$\ref{onestep}. $-$Step $3$ represents using softstair activation for step computation (Eqn. \ref{lab:step}) from location attention instead of the proposed sigmoid activation in step $3$ change of OneStep ($\S$\ref{onestep}). $-$Sigmoid represents removing the activation function from Eqn. \ref{lab:step} altogether. As the ablation shows both of our proposed changes are important to succeed in most of the tasks. Interestingly, we find here that having no activation at all in Eqn. \ref{lab:step} generally serves us better than having softstair activation. Besides that, the ablation results support our original motivation for proposing Step $2$ and Step $3$ in OneStep attention. We show several more ablation tests in Appendix \ref{ablations}.

\textbf{Additional Tasks:} In Appendix \ref{newtasks}, we introduce and explore two additional tasks - \textbf{DeDupe} and \textbf{PosRetrieve}.

\textbf{Alternate Evaluation:} In Appendix \ref{alternate_eval} we evaluate the models on edit distance instead of exact match accuracy. Edit distance serves as a more fine-grained evaluation.

\textbf{Examples:} In Appendix \ref{examples} we present some example failure cases of OneStep attention and monotonic attention.

\section{Limitations}
First, although OneStepAttn and MonoAttn perform better than LocAttn in general, they are also more restricted. Nevertheless, OneStepAttn and MonoAttn show the potential of the LocAttn-like framework with restrained room for error accumulation and slightly stronger inductive biases. Ideally, we want to improve upon them in the future to get both higher flexibility and good performance. Moreover, when building specific modelling capacities (say attending to absolute positions), we should also consider building appropriate synthetic tasks for sanity checking in a similar spirit as done in this paper. In Appendix \ref{newtasks}, we propose PosRetrieve which can be a sanity check for absolute position attention capability for future developments.  

Second, our experiments are limited to mainly synthetic tasks most of which require purely location-based attention\footnote{Although, we should note that despite their simplicity, the tasks still have been difficult to solve perfectly \cite{dubois-etal-2020-location, dehghani2018universal, Liang2021OutofDistributionGW}} but no complex synergy between content-based attention and position-based attention. More synthetic tasks for sanity checking such capacities can be built.  

Third, our exploration is currently limited to RNN-based seq2seq models. One reason for focusing on RNNs is because vanilla non-pretrained Transformers encoders can struggle to solve tasks like lookup table for decoder to do its job without specific changes \cite{csordas2022the}. Moreover, integration of location attention into Transformers is complicated by the fact that they use multiple layers of cross-attention in each timestep introducing additional variables to consider (the problem is not that our methods cannot be integrated with Transformers but that there are many ways to do so). Given these added variables, we leave investigations with Transformers for future work. 


\section{Conclusion}
We introduce several new probing tasks - ReCopy and its variants (some others in Appendix \ref{newtasks}) to enable additional diagnoses of length generalization performance of neural models. Although our proposed tasks are simple, this very simplicity can allow better isolation of failure cases and provide sanity checks for locational reasoning skills. Moreover, the new tasks are still challenging enough that none of the models explored here succeed in all of them. 

We propose a way to softly switch between the forward encodings and its reversed version to get near perfect performance in reverse variants of copy and lookup tasks that have been previously challenging to solve. We illuminate the limits of location attention and show how certain modifications in the form of OneStep attention and monotonic attention can bring massive improvement. Although, the modifications bring stronger inductive biases than location attention, they can still simulate windowed relative attention and empirically demonstrate more stable performance across datasets including more realistic ones like CFQ. 


Monotonic attention or OneStep attention can also be more broadly applicable in any context requiring \textit{list traversal} i.e. monotonic traversal through a list of items in a backpropagation-friendly manner --- for example, one application can be skill selection with a dynamic time horizon instead of a fixed one \cite{garg2022lisa}. OneStep attention is suitable if the only relevant choice during the traversal is to either stay at a position or move to the next position by a single step. Monotonic attention is suitable if we also want to allow the model to skip positions during traversal.

\section*{Acknowledgments}
This research is supported in part by NSF CAREER award \#1802358, NSF IIS award \#2107518, and UIC Discovery Partners Institute (DPI) award. Any opinions, findings, and conclusions expressed here are those of the authors and do not necessarily reflect the views of NSF or DPI. We thank our anonymous reviewers for their constructive feedback.

\bibliography{main}
\bibliographystyle{icml2023}

\newpage
\appendix

\begin{table*}[t]
\small
\centering
\def\arraystretch{1.2}
\begin{tabular}{  l | l | l} 
\hline
\textbf{Task} & \textbf{Input} & \textbf{Output}\\
\hline
DeDupe & $4$  $4$  $4$ $7$  $7$  $7$ $7$  $9$  $9$  $9$ $9$  $9$  $8$  $8$ $8$  $8$ $8$ & $4$ $7$ $9$ $8$\\
PosRetrieve & $5$ $4$ $2$ $7$ $9$ $6$ $9$ $5$ $7$ $3$ & $5$:$6$; $4$:$9$; $2$:$2$; $7$:$5$; $9$:$3$; $6$:$9$; $9$:$3$; $5$:$6$; $7$:$5$; $3$:$7$;\\
\hline 
\end{tabular}
\caption{Input-output examples for DeDupe and PosRetrieve}
\label{table:task-examples2}
\end{table*}

 \begin{table*}[t]
\small
\centering
\def\arraystretch{1.2}
\begin{tabular}{  l | l l l l | l l l l} 
\hline
\textbf{Model} & \multicolumn{4}{c}{\textbf{DeDupe}} & \multicolumn{4}{|c}{\textbf{PosRetrieve}}\\
(Length Splits)  & IID & $15$ & $30$ & $100$ & IID & $15$ & $30$ & $100$\\
\hline
Content &  $98.7$ & $37.1$ & $0$ & $0$ & $\mathbf{100}$ & $0$ & $0$ & $0$\\
Bi-Relative & $\mathbf{99.9}$ & $68$ & $0$ & $0$ & $\mathbf{100}$ & $0$ & $0$ & $0$\\
LocAttn & $99.6$ & $96.9$ & $71.4$ & $0$ & $96.3$ & $0$ & $0$ & $0$\\
OneStepAttn & $94.1$ & $51.$ & $0$ & $0$ & $70.6$ & $0$ & $0$ & $0$\\
MonoAttn & $99.7$ & $\mathbf{98.3}$ & $\mathbf{94.3}$ & $\mathbf{75}$ &  $79$ & $0$ & $0$ & $0$\\
\hline 
\end{tabular}
\caption{Accuracy of the models on different length generalization splits in DeDupe and PosRetrieve. We present the median of five runs on different seeds. We bold the best results.}.
\label{table:diagnostics3}
\vspace{-4mm}
\end{table*}

\begin{table*}[t]
\small
\centering
\def\arraystretch{1.2}
\begin{tabular}{  l | l l l | l l l | l l l | l l l} 
\hline
\textbf{Model} & \multicolumn{3}{c}{\textbf{Copy}} & \multicolumn{3}{|c}{\textbf{Reverse Copy}} & \multicolumn{3}{|c}{\textbf{Lookup}} & \multicolumn{3}{|c}{\textbf{Reverse Lookup}}\\
(Length Splits) & $15$ & $30$ & $100$ & $15$ & $30$ & $100$ & $7$ & $9$ & $11$ & $7$ & $9$ & $11$\\
\hline
Bi-ROPE & $\mathbf{100}$ & $\mathbf{100}$ & $0$ & $\mathbf{100}$ & $\mathbf{100}$ & $0$ & $\mathbf{100}$ & $\mathbf{100}$ & $\mathbf{100}$ & $89.5$ & $1.6$ & $0.2$ \\
\hline
Mix LocAttn & $99.8$ & $0$ & $0$ & $1.5$ & $0$ & $0$ & $8.6$ & $0$ & $0$ & $7.5$ & $0$ & $0$\\
LocAttn S & $\mathbf{100}$ & $26.6$ & $0$ & $\mathbf{100}$ & $74.1$ & $0$ & $\mathbf{100}$ & $46.3$ & $0$ & $99.6$ & $0.1$ & $0$\\
Mix LocAttn S & $99.7$ & $0$ & $0$ & $85.9$ & $0$ & $0$ & $33.5$ & $0$ & $0$ & $3.5$ & $0$ & $0$\\
Mix LocAttn S PR & $83.6$ & $0$ & $0$ & $99.7$ & $10.5$ & $0$ & $32.9$ & $0$ & $0$ & $1.1$ & $0$ & $0$\\
\hline
Mix OneStepAttn & $2.6$ & $0$ & $0$ & $0.4$ & $0$ & $0$ & $\mathbf{100}$ & $71.7$ & $1.6$ & $\mathbf{100}$ & $\mathbf{79.1}$ & $\mathbf{8.1}$\\
Mix OneStepAttn PR & $\mathbf{100}$ & $\mathbf{100}$ & $\mathbf{100}$ & $\mathbf{100}$ & $\mathbf{100}$ & $\mathbf{100}$ & $\mathbf{100}$ & $\mathbf{100}$ & $\mathbf{47.8}$ & $57.5$ & $0$ & $0$\\
\hline
Mix MonoAttn & $99.95$ & $5$ & $0$ & $\mathbf{100}$ & $50.8$ & $0$ &  $94.5$ & $0$ & $0$ & $53.5$ & $0$ & $0$ \\
Mix MonoAttn PR & $\mathbf{100}$ & $99.2$ & $86.7$ & $\mathbf{100}$ & $\mathbf{100}$ & $99.7$ &  $15.3$ & $0$ & $0$ & $0$ & $0$ & $0$ \\
\hline 
RMonoAttn & $\mathbf{100}$ & $\mathbf{100}$ & $\mathbf{100}$ & $\mathbf{100}$ & $\mathbf{100}$ & $99.9$ & $64.3$ & $0$ & $0$ & $72.8$ & $0$ & $0$\\
Mix RMonoAttn & $31.1$ & $0$ & $0$ & $\mathbf{100}$ & $66.3$ & $0$ & $99.8$ & $0$ & $0$ & $41.4$ & $0$ & $0$\\
Mix RMonoAttn PR & $0$ & $0$ & $0$ & $\mathbf{100}$ & $\mathbf{100}$ & $\mathbf{100}$ & $39.2$ & $0$ & $0$ & $4.3$ & $0$ & $0$\\
\hline

\textbf{Model} & \multicolumn{3}{c}{\textbf{ReCopy}} & \multicolumn{3}{|c}{\textbf{Reverse ReCopy}} & \multicolumn{3}{|c}{\textbf{Inv ReCopy}} & \multicolumn{3}{|c}{\textbf{Inv Reverse ReCopy}}\\
(Length Splits)  & $15$ & $30$ & $100$ & $15$ & $30$ & $100$ & $15$ & $30$ & $100$ & $15$ & $30$ & $100$\\
 \hline
 Bi-ROPE & $43.4$ & $0$ & $0$ & $39$ & $0$ & $0$ & $56.0$ & $0$ & $0$ & $53.1$ & $0$ & $0$\\
 \hline
Mix LocAttn & $36.1$ & $0$ & $0$ & $30.4$ & $0$ & $0$ & $99.6$ & $65.1$ & $0$ & $98.6$ & $24.1$ & $0$\\
LocAttn S &  $98.7$ & $5.2$ & $0$ & $99.9$ & $1.4$ & $0$ & $100$ & $99.3$ & $91.1$ & $99.9$ & $99.3$ & $91.8$\\
Mix Location S & $99.7$ & $0$ & $0$ & $99.6$ & $0$ & $0$ & $98.9$ & $66.6$ & $0$ & $98.8$ & $57.7$ & $0$\\
Mix Location S PR & $99.6$ & $0.4$ & $0$ & $\mathbf{100}$ & $61.4$ & $0$ & $99.8$ & $98.6$ & $88$ & $99.6$ & $98.1$ & $84.8$\\
\hline
Mix OneStepAttn & $43.5$ & $0$ & $0$ & $87.1$ & $0$ & $0$ & $10.4$ & $0$ & $0$ & $5.45$ & $0$ & $0$\\
Mix OneStepAttn PR & $99.9$ & $88.4$ & $30.2$ & $\mathbf{100}$ & $\mathbf{100}$ & $\mathbf{100}$ & $0$ & $0$ & $0$ & $0$ & $0$ & $0$\\
\hline
Mix MonoAttn & $7.09$ & $0$ & $0$ & $99.85$ & $0$ & $0$ &  $99.8$ & $82.4$ & $0$ & $\mathbf{100}$ & $82.2$ & $0$ \\
Mix MonoAttn PR & $\mathbf{100}$ & $\mathbf{100}$ & $\mathbf{100}$ & $\mathbf{100}$ & $53$ & $0.5$ & $0$ & $0$ & $0$ & $89.6$ & $77.6$ & $42.1$\\
\hline 
RMonoAttn & $\mathbf{100}$ & $\mathbf{100}$ & $99.9$ & $0$ & $0$ & $0$ & $\mathbf{100}$ & $\mathbf{100}$ & $\mathbf{99.1}$ & $\mathbf{100}$ & $\mathbf{100}$ & $\mathbf{99}$\\
Mix RMonoAttn & $98.1$ & $0$ & $0$ & $23.4$ & $0$ & $0$ &  $99.85$ & $70.5$ & $0$ & $100$ & $83.8$ & $0$\\
Mix RMonoAttn PR & $\mathbf{100}$ & $90.8$ & $51.2$ & $\mathbf{100}$ & $\mathbf{100}$ & $\mathbf{100}$ & $0$ & $0$ & $0$ & $40$ & $0$ & $0$\\
\hline
\end{tabular}
\caption{Accuracy of the models on different length generalization splits in different algorithmic diagnostic/probing tasks. We present the median of five runs on different seeds. We bold the best results.}
\label{table:diagnostics_appendix}
\vspace{-4mm}
\end{table*}

\begin{table}[t]
\small
\centering
\def\arraystretch{1.2}
\begin{tabular}{  l | l } 
\hline
\textbf{Model} & \textbf{SCAN (Length Split)}\\
 \hline
 Bi-ROPE & $10.46 \pm 3.78$ \\
 \hline
LocAttn & $24.56 \pm 17.51$\\
LocAttn S & $25.12 \pm 6.30$\\
Mix LocAttn S & $27.55 \pm 10.10$\\
Mix LocAttn S PR & $19.8 \pm 3.26$\\
\hline
OneStepAttn & $15.38 \pm 0.42$\\
Mix OneStepAttn PR & $17.67 \pm 3.54$\\
\hline
MonoAttn & $14.98 \pm 0.15$\\
Mix MonoAttn PR &  $26.68 \pm 10.27$\\
\hline
RMonoAttn & $15.28 \pm 1.47$\\
Mix RMonoAttn &  $22.92 \pm 6.39$\\
Mix RMonoAttn PR & $\mathbf{27.99 \pm 11.26}$\\
\hline 
\end{tabular}
\caption{Accuracy on SCAN length split. We report the mean and std of $5$ runs. We bold the best results.}
\label{table:scan_appendix}
\vspace{-4mm}
\end{table}

\section{Additional Tasks}
\label{newtasks}
We introduce and analyze two additional tasks here. We present examples for each in Table \ref{table:task-examples2}. 

\textbf{DeDupe:} As the name suggests, DeDupe is essentially a de-duplication task of removing contiguous repetitions\footnote{We thank one of our reviewers for this idea.} as shown in Table \ref{table:task-examples2}. The task is very similar to Inv ReCopy but with a few differences. In Inv ReCopy, repetition occurs based on a fixed rule, each number will be repeated some fixed number of times dependent on that specific number. The task in Inv ReCopy is to remove the fixed number of number-specific repetitions rather than all contiguous repetitions. For instance, if the input is ``$4$ $4$ $4$ $4$ $4$ $4$" the output for Inv ReCopy will be ``$4$ $4$" (because ``$4$" is bound to repeat three times according to the rules of ReCopy) but for the same input, the output for DeDupe will be just ``$4$". Unlike Inv ReCopy, the DeDupe function is not invertible. Thus the task cannot be inverted (whereas ReCopy is the inversion of Inv ReCopy). We generate the splits for this task (DeDupe) in the same way as the Copy task.   

\textbf{PosRetrieve:} The task of PosRetrieve is to treat the input values as position indices to retrieve from. Given an input $x$ in a list format, $x = [i, j]$, the output $y$ will be $y = i : f(i,x) ; j :f(j,x);$. Here, $f(i,x) = x[i]$ if $len(x) > i$  else $f(i,x) = n/a$. We generate the splits for this task in the same way as the Copy task. This task can more explicitly check for a models ability to choose some $p^{th}$ position item from the beginning. This was one ability for which location attention was motivated \cite{dubois-etal-2020-location}, but there was no benchmark to explicitly check for this.

We mainly focus on the forward variations of the task here but reverse variants can also be created. 

\textbf{Results:} In Table \ref{table:diagnostics3}, we show the results of the main models on DeDupe and PosRetrieve. Only MonoAttn performs decently in DeDupe which makes sense given that it is the only approach that does well in Inv ReCopy which is similar to DeDupe. DeDupe is, however, a bit harder than Inv ReCopy for MonoAttn because the encoder needs to encode information about total contiguous repetitions from the context so that MonoAttn can predict the right amount of steps without looking ahead (which it cannot without some kind of multi-step attention). In Inv ReCopy, the encoder does not have to encode any contextual information, since repetition happens according to a fixed context-independent rule. Thus, we see reduced performance in DeDupe compared to that in Inv ReCopy. None of the models is currently able to do well at PosRetrieve. This is, perhaps, expected for most models since they are currently lacking any explicit capability for modelling absolute positions but Location Attention still struggles with it despite theoretically having the capacity. We leave this task as an open challenge. 
\section{Ablations}
\label{ablations}
\subsection{Ablation Models and Other Alternatives}
We also show experimental results of different potential alternatives in the vicinity of location attentions and ablations of monotonic attention. We discuss the different models below. 

\textbf{Bi-ROPE}: Here, we use the same strategy as in $\S$\ref{dir_relative_attn} to create the encoder representations but then we apply a different positional encoding for modeling relative distances - rotary positional encodings (ROPE) \cite{Su2021Ro}. ROPE rotates the query and key vectors in space based on their sinusoidally encoded positions before using the query and key in a content-based attention as in $\S$\ref{content_attention}.  

\textbf{LocAttn S}: This is a simplified (S) version of location attention (without content attention mixing) where we set $\text{ref}_{t} = b_t$ for the first timestep to initialize the reference position using $b_t$ and then use $\text{ref}_{t} = pa_{t-1}$ like OneStep Attention/monotonic attention. This approach can be thought to be ``in between" the original location attention and monotonic attention. 

\textbf{Mix LocAttn S}: This is same as LocAttn S but with content attention mixing (Eqn. \ref{mixup}).

\textbf{Mix LocAttn S PR}: When mixing with content attention (Eqn. \ref{mixup}), there is an option to set $pa_{t-1}$ to track only the location-based attended position to keep as a reference point by setting $pa_{t-1} = \sum_{i=1}^s \lambda'_{t-1i} \cdot norm(i)$ where $\lambda'_{t-1i}$ is the location-only attention from the past timestep. We use the modifier PR (\textbf{P}osition Attention based \textbf{R}eference) to denote this way of setting $pa_{t-1}$. Mix LocAttn S PR extends Mix LocAttn S with PR. 

\textbf{Mix OneStepAttn PR:} This extends Mix OneStepAttn with PR.

\textbf{Mix MonoAttn PR:} This extends Mix MonoAttn with PR.

\textbf{RMonoAttn:} In RMonoAttn (Relaxed Monotonic Attention) we remove the sigmoid completely and overall simplify the step computation to: $\text{steps}_t = ReLU(f_{step}(l_t))$. 

\textbf{Mix RMonoAttn:} This is RMonoAttn with content attention mixing (Eqn. \ref{mixup}).

\textbf{Mix MonoAttn PR:} This extends Mix RMonoAttn with PR.

\subsection{Ablation Results}
In Table \ref{table:diagnostics_appendix}, we show the results of the above models in all the main paper tasks but SCAN and CFQ. Bi-ROPE performs similarly to Bi-Relative as we would expect. LocAttn S with its simplification performs better than LocAttn (Table \ref{table:diagnostics}) but still falls behind OneStep/monotonic attention. The Mix variant models tend to perform worse than the unmixed ones - this is because these tasks can be done purely based on positional reasoning and the content attention is more likely to confuse the models than help in these specific tasks. PR can help better track past locationally attended positions and thus improve the performance of the Mix models (compared to when they are used without PR). RMonoAttn tends to struggle more compared to MonoAttn demonstrating the value of gating with sigmoid-activated step prediction (eqn. \ref{sigmoid-gate}). 

In Table \ref{table:scan_appendix}, we show the results of the above models for SCAN. In SCAN, the trend reverses a bit - mix models tend to be here better than unmixed ones. We suspect that mixing with content attention is more beneficial in more sophisticated tasks (SCAN is at least relatively more sophisticated than others besides CFQ) because it adds more flexibility. PR can sometimes further help in SCAN too in some models.

\section{Alternate Evaluations}
\label{alternate_eval}
In Table \ref{table:diagnostics2}, we show the results of the main models on the probing tasks with mean edit distance\footnote{Computed using NLTK \cite{nltk}.} as the evaluation metric. This paints a more fine-grained picture of the differences between model performances.

 \begin{table*}[t]
\small
\centering
\def\arraystretch{1.2}
\begin{tabular}{  l | l l l | l l l | l l l | l l l} 
\hline
\textbf{Model} & \multicolumn{3}{c}{\textbf{Copy}} & \multicolumn{3}{|c}{\textbf{Reverse Copy}} & \multicolumn{3}{|c}{\textbf{Lookup}} & \multicolumn{3}{|c}{\textbf{Reverse Lookup}}\\
(Length Splits)  & $15$ & $30$ & $100$ & $15$ & $30$ & $100$ & $7$ & $9$ & $11$ & $7$ & $9$ & $11$\\
\hline
Content & $6.1$ & $21$ & $90.7$ & $6.8$ & $21.9$ & $91.7$ & $1.4$ & $4.6$ & $6.5$ & $1$ & $3.8$ & $5.4$\\
Relative & $\mathbf{0}$ & $\mathbf{0}$ & $\mathbf{0}$ & $4.4$ & $20.7$ & $90.2$ & $\mathbf{0}$ & $\mathbf{0}$ & $\mathbf{0}$ & $0.3$ & $4.9$ & $6.3$\\
LocAttn &  $\mathbf{0}$ & $14.2$ & $82$ & $1.7$ & $16.7$ & $84.9$ & $1.2$ & $3$ & $5.7$ & $1.3$ & $3.9$ & $6.2$\\
\hline
Ours \\
\hline
Bi-Relative & $\mathbf{0}$ & $\mathbf{0}$ & $\mathbf{0}$ & $\mathbf{0}$ & $\mathbf{0}$ & $\mathbf{0}$ & $\mathbf{0}$ & $\mathbf{0}$ & $\mathbf{0}$ & $\mathbf{0}$ & $\mathbf{0}$ & $\mathbf{0}$\\
OneStepAttn & $\mathbf{0}$ & $\mathbf{0}$ & $\mathbf{0}$ & $\mathbf{0}$ & $\mathbf{0}$ & $\mathbf{0}$ & $\mathbf{0}$ & $\mathbf{0}$ & $\mathbf{0}$ & $\mathbf{0}$ & $\mathbf{0}$ & $\mathbf{0}$\\
MonoAttn & $\mathbf{0}$ & $\mathbf{0}$ & $\mathbf{0}$ & $\mathbf{0}$ & $\mathbf{0}$ & $\mathbf{0}$ & $\mathbf{0}$ & $\mathbf{0}$ & $0.9$ & $0.7$ & $3$ & $4.9$\\
\hline 
\textbf{Model} & \multicolumn{3}{c}{\textbf{ReCopy}} & \multicolumn{3}{|c}{\textbf{Reverse ReCopy}} & \multicolumn{3}{|c}{\textbf{Inv ReCopy}} & \multicolumn{3}{|c}{\textbf{Inv Reverse ReCopy}}\\
(Length Splits)  & $15$ & $30$ & $100$ & $15$ & $30$ & $100$ & $15$ & $30$ & $100$ & $15$ & $30$ & $100$\\
 \hline
Content & $8$ & $53.4$ & $249.2$ & $8.1$ & $53.7$ & $249.3$ & $4$ & $19.7$ & $90.1$ &  $4.3$ & $20.1$ & $90.4$\\
Relative & $3.7$ & $31.5$ & $160$ &  $17.9$ & $60.8$ & $250.7$ & $0.5$ & $13.7$ & $73.8$ & $3.2$ & $18.4$ & $88.3$ \\
LocAttn & $0.8$ & $56.5$ & $233$ & $2.7$ & $39.1$ & $226.1$ & $\mathbf{0}$ & $0.5$ & $45.8$ & $\mathbf{0}$ & $7.8$ & $72.5$\\
\hline
Ours \\
\hline
Bi-Relative & $4.5$ & $32.4$ &  $159.6$ & $4.2$ & $31.9$ & $159.1$ & $0.7$ & $14.4$ & $74$ & $0.6$ & $13.9$ & $74$\\
OneStepAttn & $\mathbf{0}$ & $\mathbf{0}$ & $\mathbf{0}$ & $\mathbf{0}$ & $\mathbf{0}$ & $\mathbf{0}$ & $3$ & $16.6$ & $86$ & $3.1$ & $17.2$ & $85.7$\\
MonoAttn & $\mathbf{0}$ & $\mathbf{0}$ & $\mathbf{0}$ & $\mathbf{0}$ & $\mathbf{0}$ & $\mathbf{0}$ & $\mathbf{0}$ & $\mathbf{0}$ & $\mathbf{0}$ & $\mathbf{0}$ & $\mathbf{0}$ & $\mathbf{0}$\\
\hline 
\end{tabular}
\caption{Average edit distance (lower the better) of the models on different length generalization splits in different algorithmic diagnostic/probing tasks. We present the median of five runs on different seeds. We bold the best results.}.
\label{table:diagnostics2}
\vspace{-4mm}
\end{table*}

\begin{table*}[t]
\small
\centering
\def\arraystretch{1.2}
\begin{tabular}{ l | l | l } 
\hline
\textbf{Model} & \textbf{Task} & \textbf{Examples}\\
\hline
OneStepAttn & Inv ReCopy & \makecell[l]{\textbf{Input:} 1 2 5 5 5 5 5 5 7 7 7 7 7 2 9 9 9 9 9 9 9 9 9 9 3 7 7 7 7 7 \\9 9 9 9 9 4 4 4 5 5 5 3 3 4 4 4 0 1 1 4 4 4\\
\textbf{Oracle:} \colorbox{cyan}{1 2 5 5 7 2 9 9 3} 7 9 4 5 3 3 4 0 1 1 4 \\
\textbf{Prediction:} \colorbox{cyan}{1 2 5 5 7 2 9 9 3} 9 7 3 4 3
}\\
\hline
OneStepAttn & Inv ReCopy & \makecell[l]{\textbf{Input:} 9 9 9 9 9 2 1 1 6 6 6 6 6 6 3 7 7 7 7 7 6 6 6 8 8 8 8 8 1\\ 5 5 5 4 4 4 3 7 7 7 7 7 4 4 4 5 5 5 7 7 7 7 7 3 7 7 7 7 7\\
\textbf{Oracle:} \colorbox{cyan}{9 2 1 1 6 6 3 7 6 8 1} 5 4 3 7 4 5 7 3 7 \\
\textbf{Prediction:} \colorbox{cyan}{9 2 1 1 6 6 3 7 6 8 1} 4 5}\\
\hline
MonoAttn & Reverse Lookup & \makecell[l]{\textbf{Input:} t4 t4 t1 t6 t5 t2 t6 t3 t4 100 .\\
\textbf{Oracle:} \colorbox{cyan}{100 001 100 010} 011 010 111 111 101 110 \\
\textbf{Prediction:} \colorbox{cyan}{100 001 100 010} 110 100 \\
}\\
\hline 
MonoAttn & Reverse Lookup & \makecell[l]{\textbf{Input:} t4 t4 t2 t6 t5 t4 t2 t2 t3 110 .\\
\textbf{Oracle:} \colorbox{cyan}{110 010 011 101 110} 011 001 100 001 100\\
\textbf{Prediction:} \colorbox{cyan}{110 010 011 101 110}}\\
\hline 
\end{tabular}
\caption{Failure case examples of OneStepAttn and MonoAttn. We highlight the matching subsequence in cyan.}
\label{table:fail-examples}
\vspace{-4mm}
\end{table*}

\section{Examples}
\label{examples}
In Table \ref{table:fail-examples}, we show some failure case examples from OneStep attention and monotonic attention. Generally we find that they can generate the sequence correctly from the beginning up to a point (the correct generated part is highlighted in cyan) after which things go awry. 

\section{Experimental Details}
\label{exp_details}

\subsection{Architecture}
For the encoder, a stack of bi-directional GRUs is used. $e_{cls}$ is the concatenation of the final hidden state of the forward encoder GRU and the first hidden state of the backward encoder GRU. In every timestep $t$, the decoder state $h_{t-1}$ ($h_0$ initialized with $e_{cls}$) is used as a query and the encoded representations ($e_{1:s}$ or $e_{1:s}^{dir}$ depending on need) are used as keys and values. The values are passed through an extra non-linear transformation with LeakyRELU following the code from \cite{dubois-etal-2020-location} before the standard linear transformation. For the non-linear transformation we use the same $d$ neurons per layer. The output of the attention is concatenated with the embedding of the last generated token (initially some special token ``go" indicating start of sequence). The concatenation is used as input to a decoder stack of GRUs with $h_{t-1}$ as the hidden memory. The output of decoder GRU is $h_{t}$. A linear transformation over $h_{t}$ is used to change the size of the vector from $d$ (size of $h_t$) to $d_e$ (where $d_e$ is the embedding size). We then use the transpose of the embedding matrix to create the distribution over vocabulary. We select the maximum scoring token as the generated token for the timestep $t$. 

\subsection{Hyperparameters}
We attempted to keep the hyperparameters similar to \cite{dubois-etal-2020-location}. We use $64$ as the embedding size (i.e $d_e=64$) and single-layered GRUs for the encoder/decoder. The total hidden size for the encoder/decoder GRU is $128$ (therefore $d=128$). We only use one head for the attention mechanism. We use a dropout of $50\%$ on the encodings similar to \citet{dubois-etal-2020-location}. We set $\beta=5$. Following tradition we keep attention head dimension as $d/\text{heads}$ which in our case is $128$ since $\text{heads}=1$. For CFQ, we use two layered GRUs for the encoder/decoder and twice the hidden size/embedding size than above. Generally, we use a batch size $32$, a learning rate of $1e-3$ with Adam (default parameters) and no weight decay. We halve the learning rate if the accuracy plateaus for four contiguous epochs. We run the models for a maximum of $100$ epochs with $50$ patience for early stopping. More hyperparameter details can be found in the codebase. 

\section{Connections to Other Models}
Neural Turing Machines \cite{graves2014ntm} include one of the earliest incorporation and motivation of a location-based attention (distinguished from content based attention) within a modern neural network-based paradigm. \citet{luong-etal-2015-effective} proposed a Gaussian distribution-based attention for localized focus which is similar to how eccentricity effect is modeled here. Location attention from \citet{dubois-etal-2020-location} expands upon many of these prior ideas. 

In the context of Transformers, countless works proposed ways to include some form of position-based attention bias \cite{shaw-etal-2018-self, yang-etal-2018-modeling, dai-etal-2019-transformer, Wang2020Encoding, ke2021rethinking, Su2021Ro, luo2021stable, qu-etal-2021-explore, chang-etal-2021-convolutions, wu-etal-2021-da, wennberg-henter-2021-case, likhomenko2021cape, duffer2022pi, luo2022your,sun2022length} (interalia). Dynamic convolution \cite{wu2018pay} and other similar models can also be treated as forms of location attention (query-to-distance-based attention within a local window). Most of the approaches mentioned in this paragraph, however, involve emphasis on relative distances. As we find in our investigations based on two popular representatives of such forms of attention --- Relative attention \cite{dai-etal-2019-transformer} and ROPE \cite{Su2021Ro} --- they tend to struggle on tasks like ReCopy where the ideal relative distance of attention position can be arbitrarily big and can vary with every timestep.

\citet{press2022train} introduced a new positional encoding technique for transformer-based decoders for better length generalization on language modeling. They add a bias towards locality for that purpose. However, the advantage of locality bias in the contexts of seq2seq tasks that we consider here are less clear given that the ideal position of attention can be arbitrarily distant from the current timestep of decoding in our tasks. Transformers can iteratively transfer information depth-wise  through local operations but that will be also limited by the maximum layer depth. However, allowing adaptive layers \cite{Schmidhuber2012delimiting, graves2016adaptive, bai2019deq, banio2021ponder} or intermediate computation based on scratchpad \cite{maxwell2021show} may mitigate these issues. Recently, \citet{anil2022exploring} showed that length generalization in large language models can be enhanced by a careful synergy of different techniques such as scratchpad and in-context examples. 

Neural GPU \cite{Kaiser2016neuralgpu} also achieves strong length generalization performance on several algorithmic tasks but with curriculum learning. \citet{csordas2022the} solve the lookup table tasks in both directions with gated Transformer-based models but it does so only at the level of encoding (which can be already done with a Bidirectional RNN as shown in the same paper) where it only has to learn to compute and output the final function output. It does not tackle the challenge of doing the task at a seq2seq level (or in the style of a language model) which requires printing a sequence of intermediate function outputs in a rule-based manner in addition to the final output.


\end{document}